\documentclass{article}

\usepackage{algorithm}
\usepackage{algorithmic}

\usepackage{microtype}
\usepackage{graphicx}
\usepackage{subfigure}
\usepackage{booktabs} % for professional tables
\usepackage{amsmath}  
\usepackage{amsfonts} 
\usepackage{hyperref}

\usepackage[accepted]{icml2025}

\usepackage{amsmath}
\usepackage{amssymb}
\usepackage{mathtools}
\usepackage{amsthm}

\usepackage[capitalize,noabbrev]{cleveref}

\theoremstyle{plain}

\theoremstyle{definition}

\theoremstyle{remark}

\usepackage[textsize=tiny]{todonotes}

\icmltitlerunning{Mask-Enhanced Autoregressive Prediction: Pay Less Attention to Learn More}

\begin{document}

\twocolumn[
% \icmltitle{MEAP: Mask-Enhanced Autoregressive Prediction for Large Language Model Training}

\icmltitle{Mask-Enhanced Autoregressive Prediction: Pay Less Attention to Learn More}

% \icmltitle{Pay Less Attention to Learn More: \\ Pre-training Large Language Models with Mask-Enhanced Autoregressive Prediction}

% It is OKAY to include author information, even for blind
% submissions: the style file will automatically remove it for you
% unless you've provided the [accepted] option to the icml2025
% package.

% List of affiliations: The first argument should be a (short)
% identifier you will use later to specify author affiliations
% Academic affiliations should list Department, University, City, Region, Country
% Industry affiliations should list Company, City, Region, Country

% You can specify symbols, otherwise they are numbered in order.
% Ideally, you should not use this facility. Affiliations will be numbered
% in order of appearance and this is the preferred way.
\icmlsetsymbol{equal}{*}
\begin{icmlauthorlist}
\icmlauthor{Xialie Zhuang}{equal,ucas,scitix}
\icmlauthor{Zhikai Jia}{equal,scitix}
\icmlauthor{Jianjin Li}{equal,scnu}
\icmlauthor{Zhenyu Zhang}{uta}
\icmlauthor{Li Shen}{sysu}

\icmlauthor{Zheng Cao}{scitix}
\icmlauthor{Shiwei Liu}{oxford}
\end{icmlauthorlist}

\icmlaffiliation{ucas}{School of Artificial Intelligence, University of Chinese Academy of Sciences, China}
\icmlaffiliation{scitix}{SCITIX (SGP) TECH PTE. LTD., Singapore}
\icmlaffiliation{scnu}{South China Normal University, China}
\icmlaffiliation{uta}{University of Texas at Austin, USA}
\icmlaffiliation{sysu}{Sun Yat-Sen University, China}
\icmlaffiliation{oxford}{University of Oxford, UK}

\icmlcorrespondingauthor{Zheng Cao}{zcao@scitix.ai}
\icmlcorrespondingauthor{Shiwei Liu}{shiwei.liu@maths.ox.ac.uk}

% You may provide any keywords that you
% find helpful for describing your paper; these are used to populate
% the "keywords" metadata in the PDF but will not be shown in the document
\icmlkeywords{Machine Learning, ICML}

\vskip 0.3in
]

% this must go after the closing bracket ] following \twocolumn[ ...

% This command actually creates the footnote in the first column
% listing the affiliations and the copyright notice.
% The command takes one argument, which is text to display at the start of the footnote.
% The \icmlEqualContribution command is standard text for equal contribution.
% Remove it (just {}) if you do not need this facility.

%\printAffiliationsAndNotice{}  % leave blank if no need to mention equal contribution
\printAffiliationsAndNotice{\icmlEqualContribution} % otherwise use the standard text.

\begin{abstract}
Large Language Models (LLMs) are discovered to suffer from accurately retrieving key information. To address this, we propose Mask-Enhanced Autoregressive Prediction (\textbf{MEAP}), a simple yet effective training paradigm that seamlessly integrates Masked Language Modeling (MLM) into Next-Token Prediction (NTP) to enhance the latter's in-context retrieval capabilities. Specifically, MEAP first randomly masks a small fraction of input tokens and then directly performs the standard next-token prediction autoregressive using a decoder-only Transformer. MEAP eliminates the need for bidirectional attention or encoder-decoder architectures for MLM, incurring no additional computational overhead during pre-training or inference. Intensive experiments demonstrate that MEAP substantially outperforms NTP on key information retrieval and long-context reasoning tasks, while performing on par or better on commonsense reasoning tasks. The benefits of MEAP also extend to supervised fine-tuning, where it shows remarkable advantages in lost-in-the-middle scenarios, outperforming NTP by 11.77\% percentage points. 
Our analysis indicates that MEAP’s effectiveness arises from its ability to promote more distinguishable attention scores by concentrating on a reduced set of non-masked tokens. This mechanism improves the model’s focus on task-relevant signals while mitigating the influence of peripheral context. These findings position MEAP as a promising training paradigm for large language models. Code is available at \href{https://github.com/CharlieZhuang-Code/MEAP}{https://github.com/CharlieZhuang-Code/MEAP}.

% Our analysis shows that MEAP amplifies the attention differences among tokens through masking, enabling the model to focus more effectively on useful information. 

% \textcolor{red}{The performance gain of MEAP scales significantly with training tokens, as MEAP demonstrates substantial improvements over NTP baselines on the Needle in a Haystack task.}

% Analysis of attention patterns reveals that masked tokens receive significantly reduced attention compared to unmasked tokens.
\end{abstract}

\section{Introduction}
\label{sec:intro}

\looseness=-1 Next-token prediction (NTP) \citep{radford2018improving} is the foundational training objective for many large language models (LLMs), including OpenAI's GPT series \citep{brown2020language}. NTP trains models to predict the next word (or token) in a sequence, given all preceding tokens. Its scaling efficiency and exceptional performance in text generation have established it as the dominant paradigm for state-of-the-art LLMs such as GPT-4 \citep{achiam2023gpt}, LLaMa3 \citep{dubey2024llama}, Gemini 1.5 Pro \citep{team2024gemini}, and DeepSeek-V3 \citep{liu2024deepseek}. However, recent studies highlight the limitations of NTP-based LLMs in accurately retrieving key information from context \citep{liu2024lost,kamradt2023needle,nelson2024needle}.

In contrast, masked language modeling (MLM), used in BERT \citep{devlin2018bert}, adopts a denoising objective that reconstructs masked inputs using bidirectional attention. This cloze-type nature makes MLM particularly effective for tasks requiring precise information retrieval and sentence-level understanding. However, MLM's inherent focus on reconstructing masked tokens reduces its effectiveness in tasks requiring coherent and long-form text generation \citep{wang2019bert,dong2019unified}.

While intuitively appealing, combining NTP and MLM to leverage their respective strengths remains a non-trivial challenge. MLM typically operates best within two-stack encoder-decoder architectures, and performance degrades significantly when applied to decoder-only Transformers \citep{tay2022ul2}. Efforts to integrate the two often rely on unified pre-training pipelines where multiple objectives are alternated during the pretraining process \citep{dong2019unified,tay2022ul2}. However, this multi-objective approach introduces substantial complexity to the training pipeline, making it cumbersome to scale, especially for models with billions or trillions of parameters.

\begin{figure}[t]
    \centering
    \vskip 0.2in
    \includegraphics[width=1.0\linewidth]{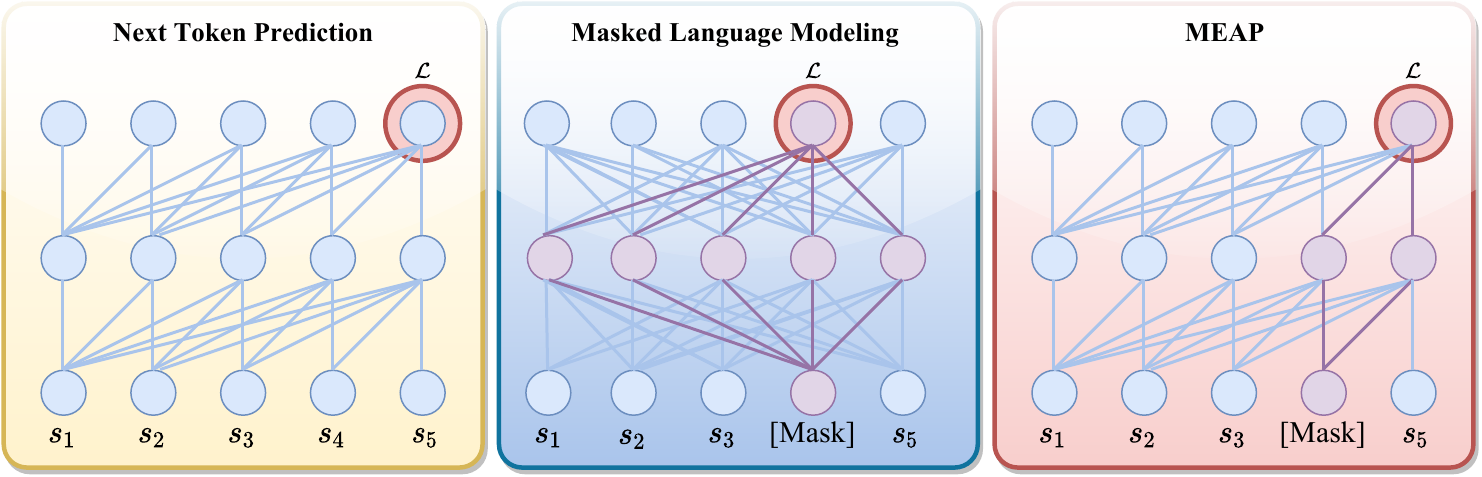}
    \vskip -0.1in
    \caption{Overview of next token prediction, masked language modeling, and our MEAP.}
    \label{fig:teaser}
    \vskip -0.2in
\end{figure}

To this end, we propose \textbf{Mask-Enhanced Autoregressive Prediction (MEAP)}, a simple yet effective LLM training paradigm that seamlessly integrates masked tokens into next-token prediction. Specifically, we first randomly mask a small fraction of the input tokens and then directly perform standard next-token prediction using a decoder-only Transformer in an autoregressive manner. This straightforward modification eliminates the need for bidirectional attention or an expensive encoder-decoder architecture, thereby incurring no additional computational overhead during training. During inference, our resulting LLMs can work as simply as LLMs that are trained with NTP with no extra engineering effort. The simplicity of MEAP enables us to enhance LLMs' performance of key information retrieval and long-context reasoning, while retaining the impressive scaling efficiency of decoder-only LLMs. Figure \ref{fig:teaser} shows the illustrations of different training paradigms.

As a general pre-training paradigm, MEAP works effectively for scenarios of pre-training and fine-tuning. For the pre-training setting, we conduct control experiments by pre-training 1.1B LLaMa-style LLMs \citep{zhang2024tinyllama} with NTP and MEAP, where the training tokens scale from 40B to 200B. Our results demonstrate that MEAP substantially improves the performance in key information retrieval tasks such as Needle in a Haystack \citep{kamradt2023needle} by up to 33\% average score and Multi-Document Question Answering (MDQA) \citep{liu2024lost} by up to 27.2 percentage points, while preserving general knowledge learned during pre-training. It is noteworthy that MEAP achieves 85.8\% accuracy with 60B training tokens on the Needle in a Haystack, while NTP requires 200B for similar performance, highlighting MEAP’s superior data efficiency in key information retrieval.
In addition, compared to the original NTP, MEAP also suffers less from hallucination.

In addition, the promise of MEAP also holds for LLM fine-tuning. Our MEAP framework
demonstrates consistent improvements across multiple commonsense reasoning tasks, achieving an average gain of 1.12 scores over the NTP baseline. On Multi-Document Question Answering, MEAP achieves an average improvement of 11.77\% across all positions.

Our analysis suggests that MEAP’s effectiveness stems from its ability to enhance attention distinguishability by focusing on a reduced set of non-masked tokens. This mechanism sharpens the model’s attention to task-relevant signals while reducing the impact of peripheral context. In essence, MEAP learns more by attending to fewer tokens.

The structure of this paper is as follows. Section \ref{sec:meap} details the MEAP algorithm. The evaluation of MEAP on LLM pre-training and fine-tuning is presented in Sections \ref{sec:eval_pre} and \ref{sec:eval_ft}, respectively. In Section \ref{sec:why}, we further analyze the underlying reasons for MEAP’s effectiveness. Section \ref{sec:ablation} provides an ablation study, and we conclude the paper in Section \ref{sec:conclusion}.

% As anticipated in Figure 1, multi-token prediction
% instructs the LLM to predict the n future tokens from each\ position in the training corpora, all at once and in parallel (Qi et al., 2020).

% Why MLM + NTP?

% Ours vs NTP loss + MLM loss

% Use a table to clarify our method compared to previous approaches, various options, no bi-directional, no encoder. 

% Benefits: (1) de-noising, (2) reducing positional bias.  

\section{Related Work}

% \subsection{Pre-training Objectives for Large Language Models} 

\textbf{Masked Language Modeling.} Pre-training is one of the most important pillars of LLMs. BERT first trained a bidirectional, encoder-only Transformer with masked language modeling (MLM), where the model is trained to predict masked input tokens. XLNet \citep{yang2019xlnet} introduced the Permutation-based Language Modeling to account for dependencies between masked tokens during training. RoBERTa \cite{liu2019roberta} further improves 
the pre-training of BERT by training the model longer,
over more data, with longer sequences, etc. MLM was further advanced by T5 \citep{roberts2019exploring}. Specifically, T5 frames every text processing task as a 'text-to-text' problem, leveraging increased lengths of corrupted tokens to achieve improved performance on classification tasks, which has contributed to its growing popularity. However, these models have shown limited performance in open-text generation and in-context learning, limiting their usage in modern LLMs.

\textbf{Next Token Prediction.} In a parallel vein, \citet{radford2019language} proposed next-token prediction (NTP) where a decoder-only Transformer is trained to predict the next token from left to right using unidirectional attention ensured by casual mask. By predicting the next token based on previously generated tokens and the given input context, NTP maintains coherence and logical flow in the generated text, well-suited for text generation. Moreover, NTP eliminates the need for an encoder, significantly improving the scalability of language models. Due to the above advantages, NTP serves as the most popular pre-training objective of modern LLMs \citep{brown2020language,achiam2023gpt,touvron2023llama,jiang2023mistral,yang2024qwen2,liu2024deepseek}. 

\textbf{Unified Training Paradigms.} There are works that propose unified training paradigms aiming to train one Transformer with multiple objective functions. For instance, UniLM \citep{dong2019unified} trains a bidirectional encoder on unidirectional language modeling (LM), bidirectional LM, and Sequence-to-Sequence LM. UL2 \citep{tay2022ul2} proposes a unified pre-training paradigm with Mixture-of-Denoisers (MoD) to combine diverse pre-training paradigms together, improving the performance over T5 and GPT. 
While effective, the preference for encoder-decoder architectures and the complicated switches among different training objectives hinder their applications in practice. 

In contrast, our approach seamlessly integrates masked tokens into NTP without incurring any additional pre-training or inference costs, while preserving the ultra-efficiency of NTP. More importantly, MEAP is more suitable for modern LLMs, as our method does not alter the core mechanism of NTP, the resulting models remain fully compatible with existing pipelines, platforms, and hardware optimized for modern LLMs. 

% \textbf{Benefits of our approach: Although MLM significantly improves the performance of a wide range of natural language understanding tasks [CITA BERT], its bidirectional nature makes it difficult to be applied to natural language generation tasks [44]. Our approach (1) does not alter the core mechanism of NTP so that the trained LLMs work as normal NTP LLMs, seamlessly compatible with existing pipelines, platforms, and hardware; (2) No extra training overhead. (3) No extra engineering effort for inference.}

\section{Mask-Enhanced Autoregressive Prediction}
\label{sec:meap}

\begin{figure}[t]
    \centering
    \resizebox{0.49\textwidth}{!}{%
        \includegraphics{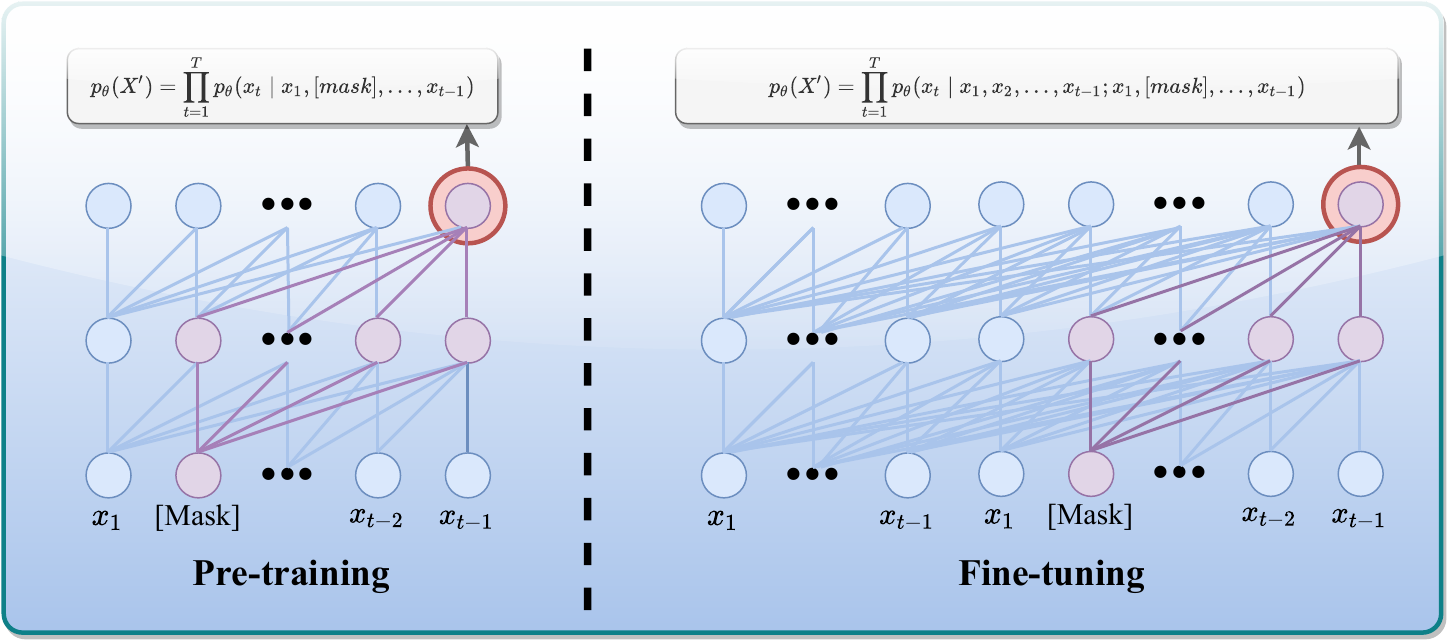}
    }
    \caption{Training frameworks of MEAP: Left (Pre-training): A certain portion of input tokens is randomly masked, followed by standard next-token prediction (NTP). Right (Fine-tuning): Training samples are duplicated, and the random masking strategy is applied to the copied sequences. Standard NTP is then performed on the modified input for fine-tuning.} 
    \label{fig:framework}
\end{figure}

In this section, we introduce Mask-Enhanced Autoregressive Prediction (MEAP).

\textbf{LLM pre-training.} To enhance the performance of LLM in handling and understanding long texts, particularly in key tasks such as key information retrieval and long context, we designed and implemented a simple yet efficient random masking strategy. The core idea of this method is to selectively mask portions of the input during the pre-training phase. Specifically, we employed a fixed proportion masking mechanism, where tokens in the input sequence are randomly masked according to a predefined percentage $P$. In this way, the model is forced to learn in the absence of some contextual information, which helps improve its deep understanding and reasoning capabilities. 

Formally, given a decoder-only Transformer $\theta$ and a sequence of input $X=(x_1, x_2, ...x_{n-1}, x_n)$, we first randomly mask a fraction of $P$ tokens having $X'=(x_1, [mask], ..., x_{t-1}, x_{t})$. We then perform the standard next-token prediction using the masked input in a left-to-right manner:

\[
p_\theta(X') = \prod_{t=1}^T p_\theta(x_t \mid x_1, [mask], \dots, x_{t-1})
\]

Same as NTP, when the model is tasked with predicting a masked token, it employs causal masked attention, using only the preceding tokens to predict the masked token. We carefully selected the masking ratio, $P=15\%$ for pre-training, to ensure that the model receives an adequate level of training difficulty and learning signals, without excessively disrupting the pre-training process. The relatively moderate number of masked tokens allows this approach to be seamlessly integrated into existing NTP frameworks, without significantly increasing pre-training overhead or altering the original training procedure.

\textbf{LLM fine-tuning.} MEAP can also be extended to fine-tuning scenarios. In this scenario, we duplicate the training samples and apply the same random masking strategy to the copied sequences during fine-tuning. The original sequences and their masked counterparts are then combined into a single input sequence to feed into the model. The cross-entropy loss is computed only with the masked tokens ($U_m$) in the answer tokens ($U_q$). This design addresses a critical concern: input sequences in supervised fine-tuning often contain key information essential for downstream tasks. Directly masking the original sequence risks removing crucial information, potentially compromising the model's performance on the target tasks. Masking the duplicated sequence incorporates MLM to NTP while avoiding this concern. We choose $P=10\%$ for fine-tuning in our experiment. We only perform MEAP for the QA pair whose answer length exceeds 50, otherwise, we perform the standard NTP for the pair. 
Formally, the following objective of MEAP for fine-tuning is:

{\small
\[
\mathcal{L}(\theta) = -\sum_{t\in U_q\cup U_m}\log p_\theta(x_t \mid x_1, \dots, x_{t-1}; \hat{x}_1, [\text{mask}], \dots,  \hat{x}_{t-1})
\]
}

% \begin{align*}
% \mathcal{L}(\theta) = - \sum_{t\in U_q\cup U_m} \log\, p_\theta(
% &x_t \mid x_1, \dots, x_{t-1}; \\
% &\hat{x}_1, [\text{mask}], \dots, \hat{x}_{t-1})
% \end{align*}
where the sequence $\{\hat{x}_i\}$ is a copy of the original sequence $\{x_i\}$ (\textit{i.e.}, $\hat{x}_i = x_i$). 

Notably, while MEAP doubles the sequence length during fine-tuning, {Figure~\ref{fig:time-inference}} shows that it achieves superior performance to NTP with only half the training time, essentially attaining stronger results with even fewer training tokens.

\looseness=-1 We believe that the effectiveness of MEAP stems from its ability to promote more distinguishable attention by focusing on fewer tokens during LLM training, as masked tokens typically receive negligible attention. This modification helps the model focus on task-relevant signals while reducing the impact of peripheral context, as verified in Section \ref{sec:why}.

% To enhance the performance of large language models in handling and understanding long texts, particularly in key tasks such as reasoning and information retrieval, we designed and implemented a simple yet efficient random masking strategy. The core idea of this method is to selectively mask portions of the input information during the model’s pretraining phase. Specifically, we employed a fixed proportion masking mechanism, where tokens in the input sequence are randomly masked according to a predefined percentage. In this way, the model is forced to learn in the absence of some contextual information, which helps improve its deep understanding and reasoning capabilities.

% Assume the input text sequence is \( X = \{x_1, x_2, \dots, x_T\} \), where \( T \) denotes the sequence length. For each input sequence, we perform random masking as follows:

% 1. Random Sampling of Mask Positions 
%    We randomly select tokens in the sequence to be masked according to a fixed proportion \( p = 15\% \). Let the set of sampled mask positions be \( \mathcal{M} \subset \{1, 2, \dots, T\} \), where \( |\mathcal{M}| = \lceil p \cdot T \rceil \).

% 2. Generation of the Masked Sequence  
%    For the tokens located at the mask positions in \( \mathcal{M} \), replace them with a special token \(\text{[MASK]}\):  
%    \[
%    X_{\text{input}} = \{x_t \mid t \notin \mathcal{M}\} \cup \{\text{[MASK]} \mid t \in \mathcal{M}\}
%    \]

\section{Experimental Results}
\label{sec:evaluation}

\begin{table*}[h]
\centering
\caption{\textbf{Pre-training Evaluation.} Zero-shot performance of MEAP and NTP on various commonsense reasoning tasks. Results are measured directly after pre-training on 200B tokens with no fine-tuning.}
\vskip 0.1in
\resizebox{0.85\textwidth}{!}{%
\begin{tabular}{@{}lccccccccc@{}}
\toprule
& \textbf{ARC-c} & \textbf{ARC-e} & \textbf{BoolQ} & \textbf{PIQA} & \textbf{HellaSwag} & \textbf{WinoGrande} & \bf OBQA & \bf Average \\
\midrule
NTP   & 22.9 & 55.7 & 53.3 & 73.6 & 44.1 & 55.0 &  23.2 & 46.2 \\
MEAP  & 25.4 & 56.4 & 59.5 & 72.3 & 43.4 & 55.3 & 22.6  & \textbf{47.8} \\
\bottomrule
\end{tabular}%
}
\label{tab:cs_pre}
\end{table*}

To evaluate the effectiveness of MEAP in training LLMs, we conduct controlled experiments comparing LLMs pre-trained/fine-tuned by MEAP with those trained by NTP.

% \textbf{Experimental Setup.} \textcolor{red}{describe the experimental setup, i.e., architecture, heads, layers, etc, training data, number of training tokens, what optimizer, following what paper.}

% \subsection{Small-scale Training}
% We first evaluate MEAP with small-scale LLMs with LLaMa architectures from 70M to 350M, following \citep{zhao2024galore}. Models are trained with the C4 dataset. 

% \subsection{Large-scale Training}
% To draw a more solid conclusion, we scale the model size to 1.1B and token size to 200B following the setting of \citet{zhang2024tinyllama}.

% \begin{table}[t]
%     \footnotesize
%     \centering
%         \caption{Performance Comparison between Masked and Non-masked Models on Various NLP Tasks}  
%     \begin{tabular}{@{}lrr@{}}
% \toprule
% Task  & Mask model & Normal model  \\
% \midrule

% ARC Challenge  & 25.4 & 22.9 \\
% ARC Easy  & 56.4 & 55.7 \\
% BoolQ  & 59.5 & 53.3 \\
% HellaSwag  & 43.4 & 44.1 \\
% PIQA  & 72.3 & 73.6 \\
% OBQA  & 22.6 & 23.2 \\
% Winogrande  & 55.3 & 55 \\
% Avg & \textbf{47.4}& \textbf{46.2} \\

%     \bottomrule
%     \end{tabular}
%     \label{tab:closedbook_and_oracle}
% \end{table}
\subsection{Pre-training Evaluation}
\label{sec:eval_pre}
\textbf{Setup.} We follow the Llama architecture \citep{touvron2023llama} as our base model. Specifically, we train 1.1B decoder-only Transformers \citep{vaswani2017attention} following the setting of \citet{zhang2024tinyllama}. Our model has 24 layers with 32 attention heads, a hidden size of 2,048, an intermediate hidden size of 5,632, and a context length of 4096. We follow the common configurations of LLM components, e.g., Rotary Positional Embedding (RoPE) \citep{su2024roformer}, Pre-Norm \citep{ba2016layer} with RMSNorm \citep{zhang2019root}, SwiGLU \citep{shazeer2020glu}, and Grouped-query Attention \citep{ainslie2023gqa}. To assess the scalability of MEAP, we increase the training token size from 40B to 60B, and further to 200B. 

For all experiments, we implement a learning rate warm-up during the first 10\% of the training steps, followed by a cosine annealing schedule, which decays the learning rate to 10\% of its initial value. 
We use the AdamW optimizer with the following settings: \( \beta_1 = 0.9 \), \( \beta_2 = 0.95 \).  The maximum learning rate is set to \( 4 \times 10^{-4} \), the minimum learning rate is \( 4 \times 10^{-5} \), and the weight decay is \( 5 \times 10^{-2} \).

\subsubsection{Language Modeling Evaluation}
% The underlying model architecture adheres to the LLaMA framework, with carefully calibrated hyperparameters: a vocabulary dimension of 32,000, embedding dimensionality of 2,048, transformer width of 5,632, attention depth of 24 layers, 32 attention heads, and per-head dimension of 64. A maximum sequence length of 4096 is employed for all models, with a batch size of 1M tokens.  

While the primary goal of MEAP is to enhance LLM performance in key information retrieval, it is essential to ensure that integrating MLM into NTP does not compromise the model’s fundamental language modeling capability. To evaluate this, we employ the LM Eval Harness benchmark \citep{eval-harness}, assessing models in a zero-shot setting. The results, presented in Table~\ref{tab:cs_pre}, show that MEAP performs comparably to, or even outperforms, NTP, achieving a 1.6\% improvement in the overall average score. This finding provides strong evidence that incorporating random masking into NTP does not degrade the model’s language modeling capacity. In the following evaluations, we will examine whether MEAP further improves performance in key information retrieval and long-context modeling.

% The results demonstrate MEAP's particular strength in reasoning tasks, with notable improvements on ARC Challenge (25.4\%) and BoolQ (59.5\%). While performance variations exist across different tasks, MEAP achieves a higher overall average score (47.4\% vs 46.2\%), suggesting that its architectural improvements effectively enhance model capabilities. These findings highlight the importance of architectural design in balancing performance across diverse language understanding tasks.

\subsubsection{Needle-in-a-Haystack Retrieval}
\label{sec:niah}
\begin{table}[h]
    % \small
    \centering
    \caption{Single needle accuracy (\%) with 32K context.}
    % \resizebox{0.3\textwidth}{!}{%
    \vskip 0.1in
    \begin{tabular}{@{}lccc@{}}
        \toprule
        \ Token & 40B & 60B & 200B \\
        \midrule
        NTP & 65.9 & 52.8 & 87.1 \\
        MEAP & \bf 80.2 & \bf 85.8 & \bf 98.2 \\
        % \midrule
        % Relative Gain & +14.3\% & +33.0\% & +11.1\% \\
        \bottomrule
    \end{tabular}
    \label{tab:niah_results}
\end{table}

\begin{figure*}[t]
  \centering
  \subfigure[NTP Pre-training]{
    \includegraphics[width=0.48\textwidth]{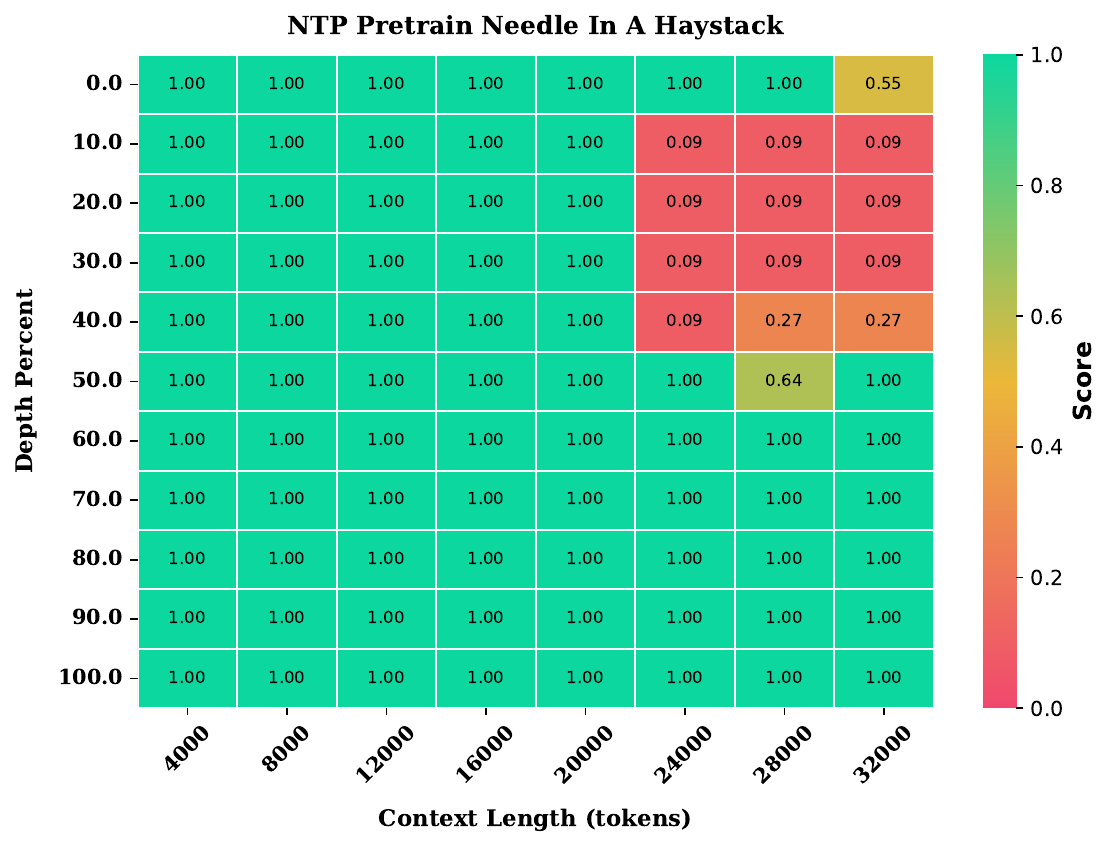}
    \label{fig:normal_pretrain}
  }
  \hfill
  \subfigure[MEAP Pre-training]{
    \includegraphics[width=0.48\textwidth]{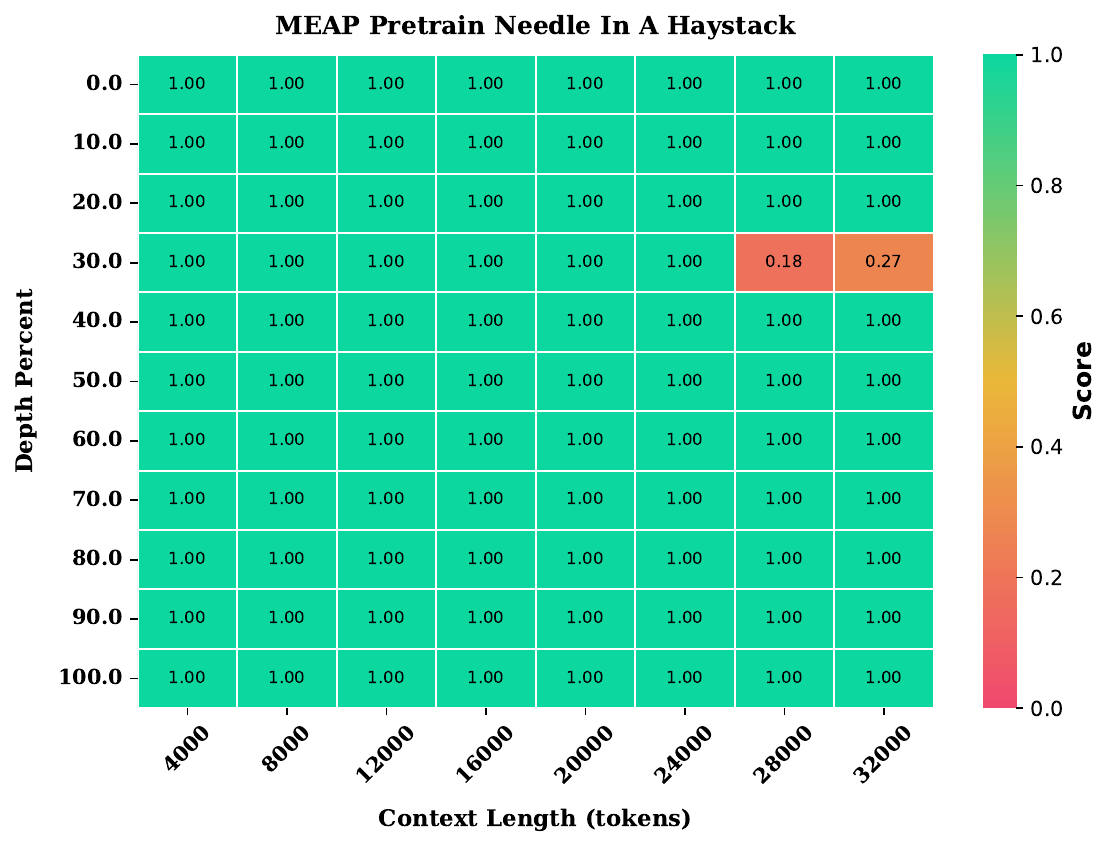}
    \label{fig:meap_pretrain}
  }
  \vspace{-1.5em}
    \caption{
        Performance comparison between NTP and MEAP on Needle In A Haystack. Scores are computed using ROUGE-1, measuring unigram overlap between model responses and expected answers.
    }
  \label{fig:32k_needle}
\end{figure*}
 
For key information retrieval, we choose the well-established Needle-in-a-Haystack evaluation \citep{liu2024lost}, where the model is asked to retrieve the random fact or statement (the `needle') in the middle of a long context window (the `haystack'). This approach provides quantitative metrics for assessing precise information extraction from extended contexts, particularly relevant for document analysis applications.

As this evaluation involves long-context modeling capacity, we follow the setting of \citet{ye2024differential} and conduct a length extension to 64K. In particular, we continue training our model for additional 4B tokens from SlimPajama \citep{sobolevaslimpajama} using the approach proposed in \citep{fu2024data}. The implementation utilizes modified Rotary Position Embeddings with $\theta_{\text{base}} = 640,000$.

To demonstrate MEAP’s scalability, we increase the training token size to 40B, 60B, and 200B, reporting the results of needle retrieval in Table~\ref{tab:niah_results}. The results show that MEAP consistently outperforms NTP across different training scales. At 40B tokens, MEAP achieves 80.2\% accuracy, significantly surpassing the baseline’s 65.9\%. The performance gap peaks at 60B tokens, with MEAP maintaining steady improvement and reaching 85.8\% accuracy. At 200B tokens, MEAP approaches optimal performance, attaining 98.2\% accuracy, while the NTP baseline still falls short of 90\% accuracy. 
It is noteworthy that MEAP achieves 85.8\% accuracy using just 60B training tokens, whereas NTP requires approximately three times as many (200B tokens) to reach a similar level. This demonstrates MEAP’s superior data efficiency over NTP in key information retrieval.

% \textcolor{red}{NTP experiences a performance drop when scaling from 40B to 60B training tokens (from 65.9\% to 52.8\%), likely due to reduced sensitivity to infrequent data points during extended training, before improving again at 200B tokens as the larger dataset provides sufficient exposure to rare patterns.}

We further illustrate the retrieval performance of our 200B-token model with a 32K context length in Figure~\ref{fig:32k_needle}. The accuracy is reported across varying answer needle depths (y-axis) and context lengths (x-axis). The results show that MEAP generally maintains perfect accuracy across different context lengths and depths, with errors limited to only two grid cells. In contrast, NTP begins to exhibit accuracy degradation at a context length of 24K, affecting a wide range of depths from 50\% to 100\%.

% The training dynamics illustrate the model's convergence characteristics. MEAP exhibits stable gradient flow throughout the training process, particularly evident in the consistent performance scaling across sequence lengths. This stability translates to improved information retrieval capabilities, as demonstrated by the quantitative metrics.

\subsubsection{Multi-Document Question Answering}

\begin{table}[h]
    \centering
    \caption{\textbf{Pre-training Evaluation.} Relative accuracy (\%) improvement of MEAP over NTP on multi-document QA.}
    \vskip 0.1in
    \label{tab:qa_improvement}
    \begin{tabular}{@{}lccccc@{}}
        \toprule
        Answer Position &  1 &  5 &  10 &  15 &20 \\
        \midrule
        10 documents & +7.6 & +7.0 & +30.6 & -- & -- \\
        20 documents & +12.4 & +4.0 & +5.1 & +3.7 & +27.2 \\
        \bottomrule
    \end{tabular}
    \vspace{-2mm}
\end{table}

We evaluate the model's ability to retrieve information from long contexts using a multi-document QA task \citep{liu2024lost} based on NaturalQuestions-Open \citep{kwiatkowski2019natural}. The task requires identifying answers from a target document while ignoring $k-1$ distractor documents, where $k$ is the total number of documents in the context. We analyze performance across two context lengths with $k=10$ and $k=20$ documents and multiple answer positions. For each query, we construct an input context containing one target document with the annotated answer and $k-1$ distractor documents that do not contain any of the answers. 64K-extended models are used for this evaluation. 

We report the accuracy improvement of MEAP over NTP in Table~\ref{tab:qa_improvement}. MEAP again consistently outperforms NTP by good margins across all configurations, with significant gains at later positions (+30.6\% at position 3 in 10-doc, +27.2\% at position 5 in 20-doc). These results indicate that MEAP enhances the model's ability to retrieve relevant information from long contexts, maintain performance across different context lengths and positions, and handle complex scenarios with multiple distractors. The improvements highlight the effectiveness of the masking strategy in enhancing the model's overall capability for long-context information retrieval tasks.

\subsubsection{Long-Context Reasoning Evaluation}
\begin{figure}[t]
\centering
\includegraphics[width=0.8\columnwidth]{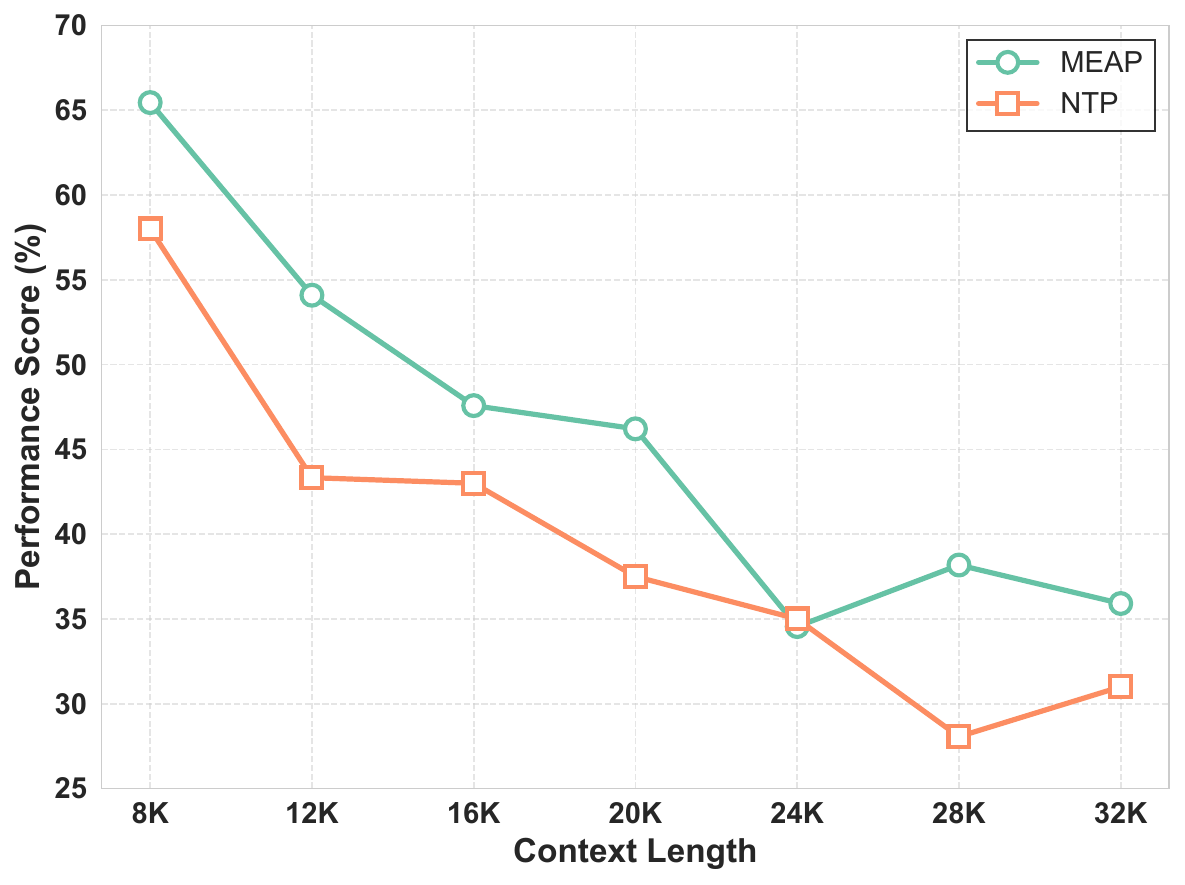}
\vspace{-1em}
\caption{Long-context reasoning performance comparison between MEAP and NTP on the Multi-Needle Reasoning Task (M-RS) across different context lengths. }
\vspace{-1.5em}
\label{fig:longinference}
\end{figure}

We evaluate long-context reasoning capabilities using the Multi-Needle Reasoning Task (M-RS) \citep{li2024needlebenchllmsretrievalreasoning}, which requires models to retrieve and extract multiple pieces of information from long texts and using them to logically answer questions that demand an integrated understanding and reasoning of various text segments.
This forces the model to distribute attention across contextually relevant tokens rather than focusing solely on local patterns.

We leverage the OpenCompass evaluation framework \citep{2023opencompass} and report the results in Figure \ref{fig:longinference}.
MEAP consistently outperforms NTP across context lengths with \textbf{6.6} percentage point average improvement. demonstrates MEAP's enhanced capacity to maintain attention coherence over extended sequences. 
% Performance degradation patterns suggest MEAP better preserves information density through dynamic masking regularization.

\subsubsection{Contextual Hallucination Evaluation}

\begin{table}[h]
\footnotesize
\vspace{-0.5em}
\caption{Accuracy (i.e., free of hallucinations) on text summarization datasets evaluated by different LLM judges.}
\vskip 0.08in
\label{tab:c_h}
\begin{center}
\begin{tabular}{lccc}
\toprule
Task & XSum & MultiNews & WikiSum \\
\midrule
NTP (Deepseek-V3) & 0.09 & 0.17 & 0.24 \\
MEAP (Deepseek-V3) & \bf 0.13 & \bf 0.19 & \bf 0.33 \\

NTP (Qwen-Plus)  & 0.16 & 0.11 & 0.21 \\
MEAP (Qwen-Plus) & \textbf{0.19}& \textbf{0.14} & \textbf{0.27} \\

NTP(GPT-4o) & 0.14 & 0.10 & 0.19 \\
MEAP(GPT-4o) & \textbf{0.16} & \textbf{0.13} & \textbf{0.24} \\
\bottomrule
\vspace{-2em}
\end{tabular}
\end{center}
\end{table}

Since MEAP improves more accurate key information retrieval, we expect it to suffer less from contextual hallucination. To verify, we evaluate MEAP in reducing contextual hallucinations on three summarization datasets: XSum \citep{narayan1808don}, WikiSum \citep{cohen2021wikisum}, and MultiNews \citep{fabbri2019multi}, following \citet{ye2024differential}. For this setting, we fine-tune the pre-trained models with Alpaca and evaluate them. We compare model-generated and reference summaries using (Deepseek-V3 \citep{liu2024deepseek}, Qwen-Plus \citep{yang2024qwen2} and GPT-4o \citep{hurst2024gpt}) as the hallucination detector across 100 random samples per dataset. As shown in Table \ref{tab:c_h}, our masking strategy achieves a consistent reduction in hallucination rates across all datasets. 

% This improvement can be attributed to the enhanced attention distribution: our masking mechanism helps the model focus on task-relevant signals while reducing the impact of peripheral context, as verified in Section \ref{sec:why}. The experimental results suggest that structured information masking during pretraining effectively improves factual consistency in abstractive summarization.

\subsection{Fine-tuning Evaluation}
\label{sec:eval_ft}

\begin{table*}[h]
\centering
\caption{\textbf{Fine-tuning Evaluation.} Performance of MEAP and NTP on various commonsense reasoning tasks. Results are measured by fine-tuning with Llama-3-8B. }
\vskip 0.1in
% \vspace{-1em}
\resizebox{0.85\textwidth}{!}{%
\begin{tabular}{@{}lccccccccc@{}}
\toprule
& \textbf{ARC-c} & \textbf{ARC-e} & \textbf{BoolQ} & \textbf{PIQA} & \textbf{HellaSwag} & \textbf{WinoGrande} & \bf OBQA & \bf Average \\
\midrule
NTP   & 53.58 & 81.10 & 83.98 & 79.27 & 62.74 & 72.06 & 39.40 & 67.30 \\
MEAP  & 55.12 & 83.21 & 83.82 & 81.01 & 63.31 & 74.27 & 38.20  & \bf 68.42 \\
\bottomrule
\end{tabular}%
}
\label{tab:cs_ft}
\end{table*}

\textbf{Setup.} We fine-tune the Llama-3-8B \citep{dubey2024llama} on the Alpaca instruction dataset \citep{alpaca}. The training configuration uses a global batch size of 512. The model is optimized with AdamW (\(\beta_1 = 0.9\), \(\beta_2 = 0.95\)), a learning rate of \(2 \times 10^{-5}\) (with 10\% warmup and cosine decay), and weight decay set to 0.01. We retain key architectural components from Llama-3, such as RoPE embeddings \citep{su2024roformer}, RMSNorm \citep{zhang2019root}, and grouped-query attention \citep{ainslie2023gqa}.

During fine-tuning, we randomly mask a portion of tokens in the assistant's response, while keeping the source context intact. Only the masked tokens are predicted during fine-tuning. The training process uses bfloat16 precision with DeepSpeed Zero Stage 2 \citep{ren2021zero}, and the Llama-3 tokenizer \citep{dubey2024llama} with a maximum sequence length of 4096 tokens.

\subsubsection{Language Modeling Evaluation}
\label{sssec:foundation}
Similar to the pre-training evaluation, we first assess MEAP's effectiveness in language modeling. Table~\ref{tab:cs_ft} presents the evaluation results. Our MEAP framework demonstrates consistent improvements across multiple tasks, achieving an average gain of 1.12 scores over the NTP baseline. The performance improvements are particularly notable on ARC-c and WinoGrande, indicating enhanced reasoning capabilities. The results highlight MEAP's effectiveness in fine-tuning complex reasoning tasks.

\subsubsection{Cross-Model Generalizability}
\label{sssec:model_sizes}

To verify that MEAP's effectiveness generalizes across different model architectures and scales, we conducted experiments on a diverse set of pre-trained LLMs. Table~\ref{tab:diff_models_cs} presents results across various commonsense reasoning tasks, while Table~\ref{tab:diff_models_mdqa} shows performance on multi-document QA tasks.

\begin{table*}[h]
\centering
\caption{\textbf{Cross-Model Generalizability.} Fine-tuning performance comparison of MEAP and NTP on commonsense reasoning tasks when applied to different architectures and model sizes.}
\vskip 0.1in
\resizebox{0.9\textwidth}{!}{%
\begin{tabular}{@{}llccccccc|c@{}}
\toprule
\textbf{Model} & \textbf{Method} & \textbf{ARC-c} & \textbf{ARC-e} & \textbf{BoolQ} & \textbf{PIQA} & \textbf{HellaSwag} & \textbf{WinoGrande} & \textbf{OBQA} & \textbf{Average} \\
\midrule
Llama-3.2-3B & NTP & 47.95 & 69.07 & 75.54 & 76.50 & 72.43 & 64.33 & 44.40 & 64.32 \\
Llama-3.2-3B & MEAP & \textbf{49.32} & \textbf{73.06} & 71.80 & \textbf{77.53} & \textbf{74.26} & \textbf{68.51} & \textbf{44.60} & \textbf{65.58} \\
\midrule
Qwen2.5-14B & NTP & 53.67 & 74.71 & 86.73 & 77.64 & 78.44 & 68.19 & 48.00 & 69.63 \\
Qwen2.5-14B & MEAP & \textbf{56.83} & \textbf{79.38} & \textbf{87.37} & \textbf{79.33} & \textbf{79.37} & \textbf{72.69} & 47.40 & \textbf{71.77} \\
\midrule
Mistral-7B-0.2 & NTP & 35.67 & \textbf{60.10} & \textbf{75.81} & 71.22 & 63.03 & \textbf{61.40} & 35.40 & 57.52 \\
Mistral-7B-0.2 & MEAP & \textbf{37.20} & 59.18 & 72.63 & \textbf{73.50} & \textbf{64.08} & 61.17 & \textbf{35.60} & \textbf{57.62} \\
\bottomrule
\end{tabular}%
}
\label{tab:diff_models_cs}
\end{table*}

The results demonstrate that MEAP consistently matches or outperforms NTP across different model architectures and sizes. On the multi-document QA task, MEAP demonstrates substantial improvements across all model architectures and sizes. These results highlight MEAP's universal effectiveness in enhancing information retrieval capabilities.

\begin{table*}[h]
\centering
\caption{\textbf{Cross-Model Generalizability.} Accuracy (\%) of MEAP and NTP on multi-document QA with 20 documents across different model architectures and sizes.}
\vskip 0.1in
\resizebox{0.9\textwidth}{!}{%
\begin{tabular}{@{}llccccc|c@{}}
\toprule
\textbf{Model} & \textbf{Method} & \textbf{Position 1} & \textbf{Position 5} & \textbf{Position 10} & \textbf{Position 15} & \textbf{Position 20} & \textbf{Average} \\
\midrule
Llama-3.2-3B & NTP & 13.60 & 12.09 & 12.54 & 12.69 & 14.35 & 13.05 \\
Llama-3.2-3B & MEAP & \textbf{23.47} & \textbf{20.34} & \textbf{20.38} & \textbf{21.96} & \textbf{23.65} & \textbf{21.96} \\
\midrule
Mistral-7B-0.2 & NTP & 36.96 & 30.55 & 27.82 & 27.55 & 38.79 & 32.33 \\
Mistral-7B-0.2 & MEAP & \textbf{37.91} & \textbf{32.98} & \textbf{31.46} & \textbf{32.22} & \textbf{43.45} & \textbf{35.60} \\
\midrule
Qwen2.5-14B & NTP & 60.00 & 51.98 & 56.01 & 56.05 & 63.39 & 57.49 \\
Qwen2.5-14B & MEAP & \textbf{61.69} & \textbf{53.71} & \textbf{57.21} & \textbf{56.65} & \textbf{66.29} & \textbf{59.11} \\
\bottomrule
\end{tabular}%
}
\label{tab:diff_models_mdqa}
\end{table*}

\subsubsection{Multi-Document Question Answering}
\label{sec:MDQA_ft}

We evaluate MEAP’s context-aware reasoning using the multi-document QA task with distractor suppression \citep{liu2024lost}. To ensure a fair comparison, we train MEAP for 2 epochs and NTP for 4 epochs, such that both approaches process a similar number of tokens. Table~\ref{tab:pos_improve} quantifies the exact match (EM) improvements across critical document positions in the 20-document setting. MEAP consistently achieves notable gains across all positions, further demonstrating its superiority over NTP. Two key patterns emerge from the experimental results:

\begin{itemize}
% \item \textit{Positional Consistency}: Sustained gains across all positions (average of a 11.77\% inprovement) indicate reduced positional bias compared to baseline's fluctuating performance (σ=4.51 vs. baseline's σ=3.32).

\item \textit{Consistent Improvement}: MEAP achieves substantial gains across all positions with an average improvement of 11.77\%, showing robust performance throughout the document range.

\item \textit{Mid-Context Advantage}: The maximum improvement at position 20 (+15.22\%) demonstrates enhanced long-range dependency modeling, crucial for connecting concepts across scientific documents.

\end{itemize}

% \item \textit{Training Efficiency}: Superior performance with 50\% fewer epochs suggests faster convergence through our dynamic masking strategy.
% \end{itemize}

These findings validate MEAP's effectiveness in preserving signal integrity across long contexts while highlighting opportunities for temporal reasoning enhancement and cross-document entity disambiguation.

\subsection{Training Efficiency Analysis}
\label{sssec:efficiency}

\begin{figure}[t]
  \centering
  \includegraphics[width=\linewidth]{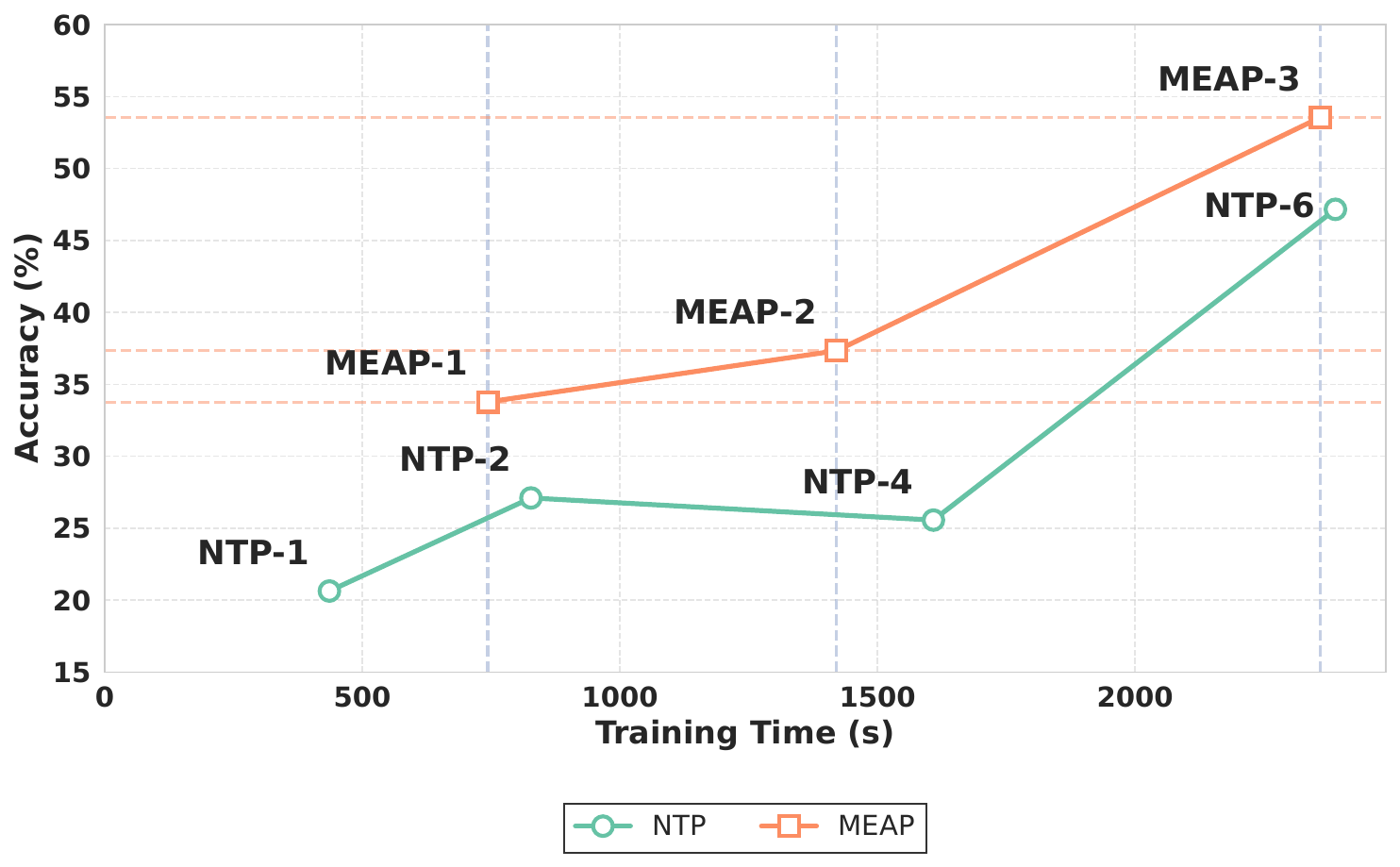}
  \caption{Comparison of fine-tuning efficiency between MEAP and NTP. `MEAP-n' refers to MEAP training for n epoch. }
  \label{fig:time-inference}
\end{figure}

MEAP introduces no additional overhead for pre-training or inference compared to standard NTP, as the only difference lies in the masking operation. During fine-tuning, MEAP requires duplicating the input sequence and training with a doubled sequence length, resulting in increased training overhead. This overhead, however, is effectively amortized by MEAP’s higher data utilization efficiency. Specifically, compared to NTP, MEAP requires only 50\% of the epochs with a similar number of tokens being processed, while outperforming the latter significantly. 

To verify, we report the results on the multi-document QA retrieval from 20 documents \citep{liu2024lost}, where retrieval performance is assessed by computing the average retrieval values across 5 positions.  
As shown in Figure~\ref{fig:time-inference}, a single epoch of MEAP training significantly outperforms two epochs of NTP training by a large margin while also reducing total training time. This highlights MEAP's data efficiency, achieving similar or better results while reducing computational resources.

\begin{table}[h]
\centering
\caption{\textbf{Fine-tuning Evaluation.} Accuracy (\%) of MEAP and NTP on multi-document QA with 20 documents.}
\vskip 0.1in
\label{tab:pos_improve}
\begin{tabular}{@{}lccccc@{}}
\toprule
\textbf{Position} & 1 & 5 & 10 & 15 & 20 \\
\midrule
NTP & 24.29 & 22.82 & 24.11 & 25.46 & 31.11 \\
MEAP  & \bf 33.22 & \bf 34.16 & \bf 36.01 &\bf 36.91 &\bf 46.33 \\
\midrule
\textbf{$\Delta$} & +8.93 & +11.34 & +11.90 & +11.45 & +15.22 \\
\bottomrule
\end{tabular}
\vspace{-3mm}
\end{table}

% We assess the training efficiency of MEAP compared to NTP on the Alpaca dataset. As shown in Figure~\ref{fig:time-inference}, MEAP achieves similar or better performance than NTP, but with significantly reduced training time. 

% A detailed time scaling analysis shows that MEAP exhibits sublinear scaling with respect to time, following the relationship $t \propto n^{0.92}$, while NTP follows a more linear scaling of $t \propto n^{1.01}$. This efficiency gain in MEAP is attributed to optimized memory access patterns and more efficient gradient computation, making it a highly effective alternative to traditional NTP methods.

In summary, MEAP delivers significant training time reductions with improved or comparable performance on the retrieval task, highlighting its efficiency and effectiveness in large-scale training scenarios.

\section{Why Does MEAP Work?} 
\label{sec:why}
In this section, we attempt to interpret the underlying reasons for the effectiveness of MEAP. We conjecture that MEAP's effectiveness stems from its ability to promote more distinguishable attention by focusing on fewer tokens during LLM training, as masked tokens \texttt{[MASK]} are expected to receive marginal attention scores. 

While effective, attention mechanisms in LLMs often struggle with long-context understanding, where redundant and non-informative attention is assigned to tokens \citep{liu2024lost,li2024long}. A plausible explanation is that the attention module relies on the Softmax function to normalize attention scores within (0, 1), which tends to minimize differences among tokens, especially when training on sequences of thousands of tokens. This bears some resemblance to \citet{martins2020sparse}'s findings on sparse attention mechanisms which implicitly relate to their core mechanism (reducing attention to tokens).

Furthermore, LLMs exhibit a phenomenon known as attention sinks, where the initial few tokens receive disproportionately high attention scores compared to the rest \citep{xiao2023efficient}. Collectively, these factors lead to small and nearly indistinguishable attention scores across tokens, which is generally undesirable. When LLMs fail to properly differentiate between tokens, they are more likely to generate incorrect outputs. 

By randomly replacing tokens with masks, MEAP implicitly penalizes the attention scores at masked positions, thereby amplifying the attention differences among non-masked tokens. This masking mechanism encourages the model to generate more distinguishable attention scores, allowing it to focus on task-relevant texts while mitigating the influence of peripheral context. We validate this hypothesis through the following experiments.

\subsection{Masking Leads to More distinguishable Attention}

\begin{figure*}[t]
  \centering
    \resizebox{0.85\textwidth}{!}{%
  \subfigure[NTP]{
    \includegraphics[width=0.48\textwidth]{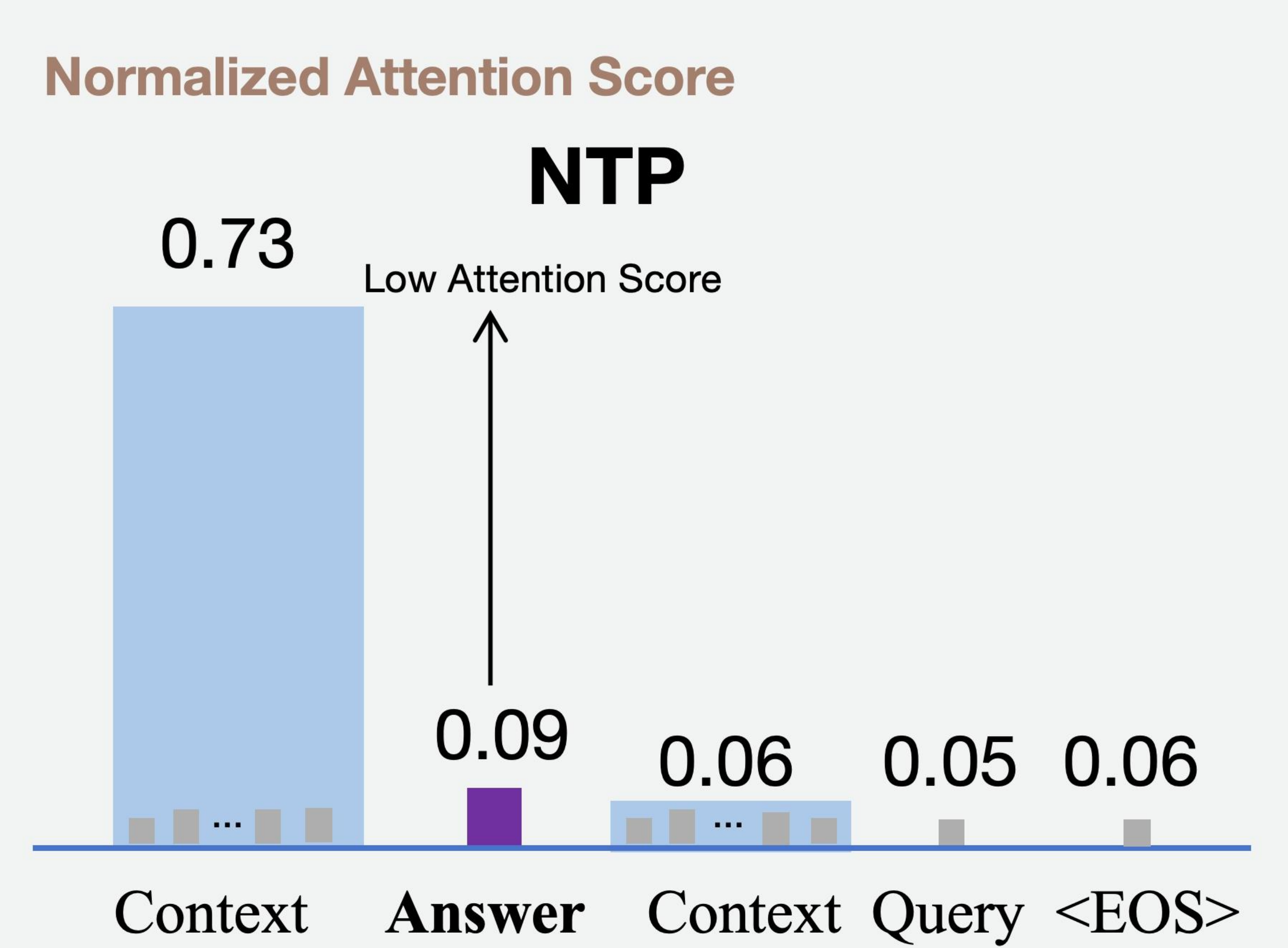}
    \label{fig:ntp_attention}
  }
  \hfill
  \hspace{5em}
  \subfigure[MEAP]{
    \includegraphics[width=0.48\textwidth]{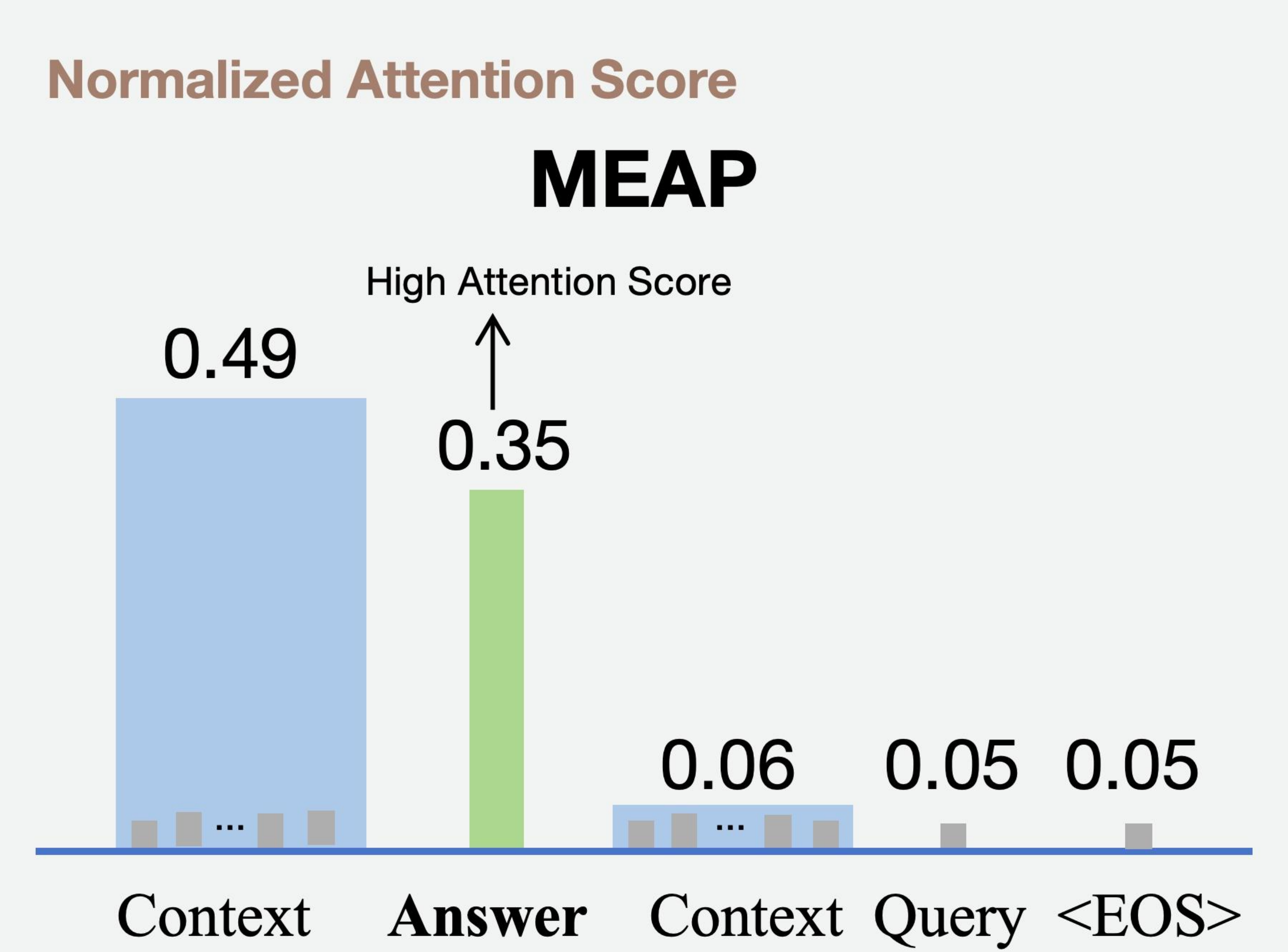}
    \label{fig:meap_attention}
  }
  }
  \vspace{-1em}
  \caption{Attention distribution comparison between NTP and MEAP during inference. The input sequence consists of: context-before (``In the heart of Paris, the Eiffel Tower stands tall, symbolizing both the city and the entire country.''), answer (``Designed by Gustave Eiffel''), context-after (``, it was completed in 1889 for the World's Fair. Originally criticized for its unusual design, it has since become one of the most recognizable landmarks in the world. Tourists from all over the globe visit it every year, making it one of the most photographed monuments.''), and query (``question: Who designed the Eiffel Tower?''). MEAP allocates a much higher attention score to answer-relevant tokens (0.345) compared to NTP (0.094).}
  \label{fig:attention_comparison}
\end{figure*}

To elucidate the mechanistic impact of our masking strategy on model behavior, we conducted a detailed analysis of attention distribution patterns. Our experimental protocol involved sampling 500 sequences. The original unmodified samples refer to the input sequence of NTP $X_{N}$, and their masked counterparts (same samples with 15\% masks), $X_{M}$, designated as the input for MEAP. These sequence pairs were then processed through our 1.1B models pre-trained with NTP and MEAP, respectively. We compare two values: (1) Attention Score Decay: the percentage decrease in the averaged attention score at masked positions, computed as:
\[
 \frac{Att(X_{N}[mask=1]) - Att(X_{M}[mask=1])}{Att(X_{N}[mask=1])}
\]
(2) Attention Variance Increase: the attention variance increase at non-mask positions, computed as:
\[
 Var(Att(X_{M}[mask=0])) - Var(Att(X_{N}[mask=0]))
\]
\textbf{Expectations.} We anticipate that the average attention score at masked positions will undergo a significant decline in the MEAP-trained model, indicating that masked tokens receive minimal attention in MEAP. Consequently, this reduction is expected to increase the attention variance at non-masked positions, making the attention distribution in MEAP more distinguishable compared to NTP.

\textbf{Results.} Table \ref{tab:attention_decay} confirms our expectations. MEAP assigns 53.34\% less attention to masked tokens, resulting in a 7.80\% increase in attention variance. This finding validates that MEAP learns more distinguishable attention scores compared to NTP.

\begin{table*}[t]
\centering
\caption{Performance comparison of different mask ratios in pre-training and fine-tuning. The best results are highlighted in bold.}
\vskip 0.1in
\label{tab:mask-ratio-results}
\begin{tabular}{l|ccccc|cccc}
\toprule
& \multicolumn{5}{|c|}{Pre-training}  & \multicolumn{4}{|c}{Fine-tuning} \\
\midrule
\textbf{Mask Ratio} & NTP & 0.05 & 0.10 &  0.15 & 0.20 & NTP & 0.05 & 0.10 & 0.15 \\
\midrule
Accuracy &0.52 & 0.54 & 0.56 & \bf 0.58 & 0.56 & 0.72 & 0.77 & \bf 0.81 & 0.71 \\ 
% \multirow{Pre-training} & 0.05 & 0.53 \\
% & 0.10 & 0.54 \\
% & \textbf{0.15} & \textbf{0.57} \\
% & 0.20 & 0.56 \\
% \midrule
% \multirow{Fine-tuning} & NTP (3 epochs) & 0.66 \\
% & NTP (6 epochs) & 0.72 \\
% & MEAP (0.05) & 0.77 \\
% & \textbf{MEAP (0.10)} & \textbf{0.81} \\
% & MEAP (0.15) & 0.64 \\
\bottomrule
\end{tabular}
\end{table*}

\subsection{MEAP Focus More on Task-Relevant Tokens}

To verify if MEAP learns more effective attention, we measure the average attention scores that the model assigns to different input segments. We structured our input sequences into distinct segments: context-before, answer, context-after, query, and EOS token. The complete input sequence was formed by concatenating these segments sequentially, followed by an EOS token. This structured format enabled precise tracking of attention allocation across different functional components. 

\textbf{Expectation.} Our expectation is that MEAP tends to amplify attention to answer spans and meanwhile reduce the attention to less relevant tokens.  

\textbf{Results.} The attention distributions during inference for both models are visualized in Figure~\ref{fig:attention_comparison}. Notably, MEAP exhibits a substantial improvement in answer-relevant attention (34.5\% vs. 9.4\%) while reducing the dominance of context-before attention from 73.1\% to 49.1\%. Both models maintain similar attention levels for peripheral components, including context-after, query sections, and EOS tokens (all approximately 5\%–6\%). These results demonstrate that the MEAP framework enhances attention allocation during inference, prioritizing key information more effectively.
% \begin{table}[h]
% \footnotesize
% \caption{Attention score comparison between NTP and MEAP.}
% \vskip 0.05in
% \label{tab:attention_decay}
% \begin{center}
% \begin{tabular}{lccc}
% \toprule
% Input Length & Mask Ratio & Score Decay & Var. Increase \\
% \midrule
% 1,024 & 0.15 & 34.08\% & 12.66\% \\
% 4,096 & 0.15 & 53.34\% & 7.80\% \\
% \bottomrule
%   \vspace{-2em}
% \end{tabular}
% \end{center}
% \end{table}

\begin{table}[h]
\footnotesize
\caption{Statistical analysis of attention score patterns between NTP and MEAP.}
\vskip 0.05in
\label{tab:attention_decay}
\begin{center}
\begin{tabular}{lccc}
\toprule
Length & Metric & Value & T-Stat/P-Value \\
\midrule
1024 & Score Decay & 34.08\% & -25.71/$<$1e-6 \\
1024 & Var. Increase & 12.66\% & 12.26/$<$1e-6 \\
\midrule
4096 & Score Decay & 53.34\% & -9.97/$<$1e-6 \\
4096 & Var. Increase & 7.80\% & 5.22/$<$1e-6 \\
\bottomrule
  \vspace{-2em}
\end{tabular}
\end{center}
\end{table}

% \section{Ablation Study and Further Investigation}
\section{Ablation Study}
\label{sec:ablation}

% \subsection{Ablation Study of Mask Ratio}
\label{sec:mask-ratio-ablation}

We conduct ablation studies on the mask ratio for both pre-training and fine-tuning settings. Table~\ref{tab:mask-ratio-results} summarizes the results. For pre-training, we evaluate our pre-trained model in Section \ref{sec:eval_pre} on the Multi-Document QA task using the \texttt{nq-open-oracle} dataset \citep{liu2024lost}. For fine-tuning, we train MEAP on the Alpaca dataset \citep{alpaca2023} for 3 epochs with different mask ratios against standard NTP baselines with 6 epochs for a fair comparison.
The results show that a mask ratio of \textbf{0.15} achieves the best performance in pre-training, while a mask ratio of \textbf{0.10} yields the highest accuracy in fine-tuning. MEAP consistently outperforms standard NTP in pre-training and fine-tuning, demonstrating its effectiveness in leveraging masked tokens for improved performance.

\subsection{Effect of Different Masking Strategies}
\label{sec:masking-strategies}
To further investigate the impact of masking patterns, we compare three distinct masking strategies: Random Masking (our default approach), 5-Span Masking (consecutive spans of 5 tokens), and 50-Span Masking (longer spans of 50 consecutive tokens). We evaluate these strategies using a 0.3B parameter model pre-trained on 5B tokens, with results presented in Table~\ref{tab:masking-strategies}.

\section{Conclusion}
\label{sec:conclusion}
This work addresses challenges in information processing through a straightforward approach that masks 10\%–15\% of input while maintaining traditional prediction methods. Our results show significant improvements in comprehension across longer contexts, achieved without additional computational costs. This approach demonstrates remarkable efficiency, matching performance metrics with just 60B training examples that typically require 200B examples with conventional methods. The results indicate that this strategy leads to more effective processing of key information through improved focus on relevant content. Since it requires no structural changes, this method can be readily integrated into existing systems without disrupting workflows.
% \clearpage
\newpage % 
\section*{Impact Statement}

This work proposes a modified pre-training paradigm that may influence how both industry and academia approach language model training. MEAP integrates seamlessly with existing LLM frameworks without requiring additional engineering effort or computational resources. While the improvement in information retrieval and reasoning capabilities could have broad implications for downstream applications, the method's computational efficiency and architectural compatibility mean it can be readily adopted within current training infrastructures. We anticipate this work will contribute to more efficient model development while maintaining established training pipelines and computational requirements.

% \begin{table}[htbp]
% \centering
% \caption{Performance Comparison between Masked and Non-masked Models on Various NLP Tasks}
% \label{tab:model-comparison}
% \begin{tabular}{|l|l|c|c|}
% \hline
% task & metric & mask 200b & nomask 200b \\
% \hline
% \multirow{2}{*}{ARC Challenge} & acc & 0.2543 & 0.2295 \\
% & acc\_norm & 0.2850 & 0.2756 \\
% \hline
% \multirow{2}{*}{ARC Easy} & acc & 0.5640 & 0.5572 \\
% & acc\_norm & 0.4962 & 0.4903 \\
% \hline
% BoolQ & acc & 0.5954 & 0.5333 \\
% \hline
% Commonsense QA & acc & 0.1974 & 0.1957 \\
% \hline
% \multirow{2}{*}{HellaSwag} & acc & 0.4338 & 0.4405 \\
% & acc\_norm & 0.5628 & 0.5741 \\
% \hline
% \multirow{2}{*}{LAMBADA (OpenAI)} & acc & 0.4648 & 0.4883 \\
% & perplexity & 12.8344 & 11.9117 \\
% \hline
% \multirow{2}{*}{LAMBADA (Standard)} & acc & 0.4044 & 0.4095 \\
% & perplexity & 19.5247 & 19.7397 \\
% \hline
% \multirow{2}{*}{PIQA} & acc & 0.7291 & 0.7356 \\
% & acc\_norm & 0.7291 & 0.7394 \\
% \hline
% Winogrande & acc & 0.5691 & 0.5501 \\
% \hline
% mmlu & acc & 0.2379 & 0.2323 \\
% \hline
% \end{tabular}
% \end{table}

% \begin{table}[htbp]
% \centering
% \caption{Contextual Hallucination Evaluation}
% \label{tab:hallucination}
% \begin{tabular}{|l|c|c|c|}
% \hline
% model & XSum & MultiNews & WikiSum \\
% \hline
% nomask 200b & 0.09 & 0.17 & 0.24 \\
% mask 200b & 0.13 & 0.19 & 0.33 \\
% \hline
% \end{tabular}
% \end{table}
% In the unusual situation where you want a paper to appear in the
% references without citing it in the main text, use \nocite

\bibliography{example_paper}
\bibliographystyle{icml2025}

%%%%%%%%%%%%%%%%%%%%%%%%%%%%%%%%%%%%%%%%%%%%%%%%%%%%%%%%%%%%%%%%%%%%%%%%%%%%%%%
%%%%%%%%%%%%%%%%%%%%%%%%%%%%%%%%%%%%%%%%%%%%%%%%%%%%%%%%%%%%%%%%%%%%%%%%%%%%%%%
% APPENDIX
%%%%%%%%%%%%%%%%%%%%%%%%%%%%%%%%%%%%%%%%%%%%%%%%%%%%%%%%%%%%%%%%%%%%%%%%%%%%%%%
%%%%%%%%%%%%%%%%%%%%%%%%%%%%%%%%%%%%%%%%%%%%%%%%%%%%%%%%%%%%%%%%%%%%%%%%%%%%%%%
\newpage
\appendix
\onecolumn

\section{Experimental Details of Pre-training}

\subsection{Architecture and Hyperparameters}

This section outlines the pre-training hyperparameters of the MEAP model, designed to ensure efficient training and optimal performance. The sequence length is fixed at 4096 tokens, enabling the model to handle long-range dependencies while maintaining computational efficiency. The learning rate schedule includes an initial warm-up phase for the first 10\% of training steps, followed by cosine decay to 10\% of the initial value, allowing gradual and precise parameter adjustments. The AdamW optimizer is used with standard hyperparameters $\beta_1 = 0.9$ and $\beta_2 = 0.95$ to stabilize the optimization process. Learning rate bounds are set between $4 \times 10^{-4}$ and $4 \times 10^{-5}$ to ensure effective learning throughout training, while a weight decay of $5 \times 10^{-2}$ helps prevent overfitting and promote generalization by penalizing excessively large weights. Complete training hyperparameters are documented in Table~\ref{tab:hyperparameters}.

The model sizes and corresponding hyperparameters are shown in Table \ref{tab:model_hyperparameters}.

\begin{table*}[h]
    \caption{Hyperparameters of training}
    \label{tab:hyperparameters}
    \vskip 0.1in
    \begin{center}
    \begin{tabular}{cc}
    \toprule
Name & Hyperparameter \\
\midrule
optimizer & AdamW \\
lr schedule & cosine \\
clip & 1.0 \\
max learning rate & $4 \times 10^{-4}$ \\
min learning rate & $4 \times 10^{-5}$ \\
weight\_decay & $5 \times 10^{-2}$ \\
sequence length & 4096 \\
batch size & 256 \\
epoch & 1\\
    \bottomrule
    \end{tabular}
    \end{center}
    \vskip -0.1in
\end{table*}

\begin{table*}[h]
    \caption{Hyperparameters of pretrained MEAP models. Data amount are specified in tokens.}
    \label{tab:model_hyperparameters}
    \vskip 0.1in
    \begin{center}
    \begin{tabular}{cccccccc}
    \toprule
    Params & Hidden & Intermediate & Heads & Layers & Steps & Data amount \\
    \midrule
    100M & 768 & 2048 & 12 & 12  & 2K & $2 \mathrm{~B}$ \\
    500M & 1024 & 4864 & 16 & 24  & 10K & $10 \mathrm{~B}$ \\
    $1.1 \mathrm{~B}$ & 2048 & 5632 & 24 & 32 & 190K & $200 \mathrm{~B}$ \\
    \bottomrule
    \end{tabular}
    \end{center}
    \vskip -0.1in
\end{table*}

\subsection{Pre-training Loss of Difference Model Sizes}
\begin{figure}[t]
    \centering
    \vskip 0.1in
    \includegraphics[width=1.0\linewidth]{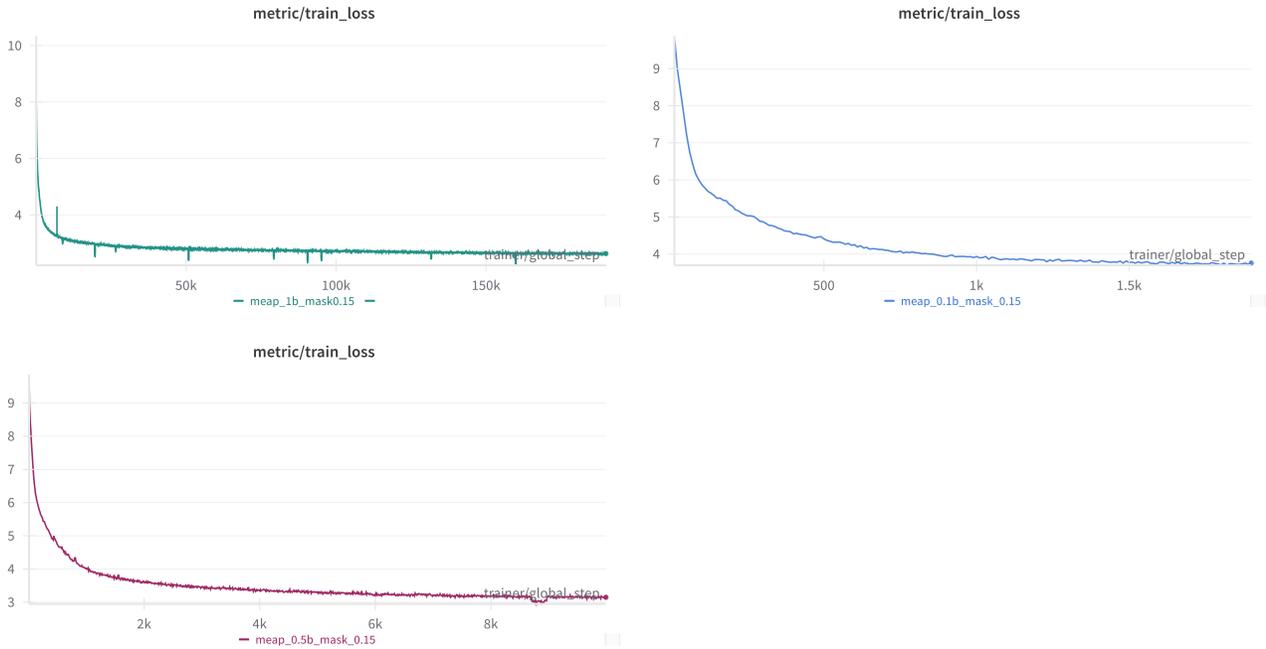}
    \vskip -0.1in
    \caption{Overview of loss for all size pretrained-models}
    \label{fig:all loss}
    \vskip -0.2in
\end{figure}

The loss curves of the MEAP model at various sizes, as shown in Figure \ref{fig:all loss}, provide a detailed visualization of the model's performance across different scales. 

 \subsection{Language Modeling Evaluation Of All Size Models for Pre-training}

As shown in Table \ref{tab: all res}, we present the evaluation results of models of different scales implemented using our method. To comprehensively assess the language modeling performance of these models, we conducted a detailed analysis for each model, with particular focus on their performance  at varying scales.

\begin{table*}[h]
    \caption{Results of all size MEAP pretrained models}
    \vskip 0.1in
    \begin{center}
    \label{tab: all res}
    \begin{tabular}{ccccc}
    \toprule
    Benchmark & 100M  & 500M & 1.1B \\
    \midrule
    ARC-Challenge & 17.32  & 18.4 & 25.4 \\
    ARC-Easy & 31.99  & 42.0 & 56.4 \\
    BoolQ & 45.14  & 55.63 & 59.5 \\
    HellaSwag & 26.82  & 30.77 & 43.4 \\
    OpenBookQA & 11.41  & 16.40 & 22.6 \\
    PIQA & 58.49  & 66.81 & 72.3 \\
    WinoGrande & 52.09  & 49.57 & 55.3 \\
    \textbf{Avg}&34.75&39.94&47.85\\
    \bottomrule
    \end{tabular}
    \end{center}
    \vskip -0.1in
\end{table*}

\subsection{Pretrained Model Evaluation Under Different Masking Rates}

As shown in Table \ref{tab:dff}, we present the evaluation results of models implemented with our approach, where different mask rates are applied during training. A comprehensive and detailed analysis of the language modeling performance is conducted for each mask rate, with a focus on how varying levels of masking influence key performance metrics. This analysis elucidates the effects of mask rates on the model’s ability to handle diverse linguistic tasks, highlighting any changes in accuracy as the mask rate is adjusted.

\begin{table*}[ht]
    \caption{Results of the 1.1b MEAP model under different masking rates }
    \vskip 0.1in
    \label{tab:dff}
    \begin{center}
    \begin{tabular}{ccccc}
    \toprule
     &Mask Ratio&Mask Ratio&Mask Ratio \\
    Benchmark & 0.15 & 0.05 & 0.1  \\
    \midrule
    ARC Challenge  & 25.4 & 26.11 & 24.3 \\

    ARC Easy & 56.4 & 56.1 & 54.3 \\
   
    BoolQ & 59.5 &56.5&53.4\\
    HellaSwag & 43.4 & 43.69  & 43.85  \\

    OpenBookQA  & 22.6 & 22.0& 21.8 \\
    PIQA &  72.3  & 72.63 & 72.91\\
    Winogrande & 55.3& 56.4 &56.91  \\
    \textbf{Avg} &47.84&47.63&46.78\\
    \bottomrule
    \end{tabular}
    \end{center}
    \vskip -0.1in
\end{table*}

\begin{table*}[t]
\centering
\caption{Performance comparison of different masking strategies against NTP baseline.}
\vskip 0.1in
\label{tab:masking-strategies}
\resizebox{0.85\textwidth}{!}{%
\begin{tabular}{lcccccccc|c}
\toprule
\textbf{Method} & \textbf{ARC-c} & \textbf{ARC-e} & \textbf{BoolQ} & \textbf{PIQA} & \textbf{HellaSwag} & \textbf{WinoGrande} & \textbf{OBQA} & \textbf{Average} & \textbf{MDQA} \\
\midrule
NTP-0.3B & 18.00 & 37.75 & 58.44 & 62.62 & 28.56 & 50.67 & 13.60 & 40.09 & 0.187 \\
\midrule
Random Mask & 21.84 & 35.44 & 61.25 & 61.04 & 29.50 & 51.46 & 27.40 & \textbf{41.13} & \textbf{0.218} \\
5-Span Mask & 21.42 & 35.61 & 60.40 & 62.08 & 29.81 & 51.07 & 27.60 & 41.14 & 0.168 \\
50-Span Mask & \textbf{23.46} & \textbf{36.20} & 59.54 & \textbf{62.84} & \textbf{30.43} & 50.99 & \textbf{28.00} & 41.64 & 0.189 \\
\bottomrule
\end{tabular}
}
\end{table*}

 \subsection{Details Of Contextual Hallucination Evaluation}

Here are the prompt for summarization generation, where "doc" is the original text to be summarized.

\fbox{
    \parbox{0.8\textwidth}{
Summarize the following article: {doc}
    }
}

We use the following prompts to let the Deepseek v3 model perform binary classification to determine whether there is hallucination in the model output compared to the human summary.

\fbox{
    \parbox{0.8\textwidth}{
The "model output" is the output of the model, and the "predicted label" is the manually annotated label. Please compare the "model output" with the "predicted label". By comparing the two, check if the "model output" is similar. If it is similar, return 1; otherwise, return 0.
An explanation of the output is required.
Here is the output format I provide. Please follow it strictly!!
Score: xx
    }
}
\\ \\ \\
 \subsection{Details of Attention Distribution of MEAP and NTP}

To validate the generality of attention changes, we conducted corresponding tests on additional examples and observed that the attention changes in these examples are consistent with the results presented in the main text. The specific changes are detailed in Table \ref{tab:ex1}, Table \ref{tab:ex2}, and Table \ref{tab:ex3}.

\begin{table*}[t]
    \caption{Attention change of example 1 }
    \vskip 0.1in
    \label{tab:ex1}
    \begin{center}
    \begin{tabular}{p{5cm}p{5cm}cc}
    \toprule
     Area&Content&MEAP attention& NTP attention \\
    \midrule
    context before&The Great Wall of China,stretching over 13,000 miles, is one of the most impressive feats of ancient engineering.&0.491&0.731\\
    \midrule
    answer&Built to protect Chinese states from invasions& 0.329&0.108\\
    \midrule
    context after& the wall took several dynasties over 2,000 years to complete. Its immense length and historical significance make it a popular tourist attraction today. The wall's construction involved countless workers, many of whom faced difficult conditions.&0.078&0.80\\
    \midrule
    query&question:What was the purpose of the Great Wall of China?&0.067&0.070\\
    \midrule
    eos& \texttt{<s>} &0.069&0.071\\
    \bottomrule
    \end{tabular}
    \end{center}
\end{table*}

\begin{table*}[t]
    \caption{Attention change of example 2 }
    \vskip 0.1in
    \label{tab:ex2}
    \begin{center}
    \begin{tabular}{p{5cm}p{5cm}cc}
    \toprule
     Area&Content&MEAP attention& NTP attention \\
    \midrule
    context before&In the early 20th century, Albert Einstein introduced his theory of relativity, which changed the way we understand space, time, and gravity.&0.435&0.694\\
    \midrule
    answer&His famous equation, E=mc²& 0.386&0.115\\
    \midrule
    context after&shows the relationship between energy and mass. Einstein’s ideas revolutionized physics, and his work led to the development of technologies like GPS and nuclear energy. Despite facing initial skepticism, his theories were eventually proven through experiments and observations, earning him a Nobel Prize in Physics in 1921.&0.066&0.074\\
    \midrule
    query&question:What famous equation did Albert Einstein create?&0.057&0.060\\
    \midrule
    eos& \texttt{<s>} &0.055&0.057\\
    \bottomrule
    \end{tabular}
    \end{center}
\end{table*}

\begin{table*}[t]
    \caption{Attention change of example 3 }
    \vskip 0.1in
    \label{tab:ex3}
    \begin{center}
    \begin{tabular}{p{5cm}p{5cm}cc}
    \toprule
     Area&Content&MEAP attention& NTP attention \\
    \midrule
    context before&At the center of Rome, the Colosseum rises as a magnificent testament to ancient Roman architecture, symbolizing the grandeur of the Roman Empire.&0.579&0.748\\
    \midrule
    answer&Constructed between 70 and 80 AD under the emperors Vespasian and Titus,& 0.219&0.065\\
    \midrule
    context after&it was used for gladiatorial contests and public spectacles. Once a symbol of Roman power, the Colosseum has weathered centuries of change but remains one of the most iconic structures in the world. Tourists flock to see it every year, making it one of the most photographed monuments in history.&0.071&0.067\\
    \midrule
    query&question:Who built the Colosseum?&0.063&0.050\\
    \midrule
    eos& \texttt{<s>} &0.068&0.069\\
    \bottomrule
    \end{tabular}
    \end{center}
\end{table*}

\clearpage

\section{Details of Fine-tuning Experiments}
\subsection{Architecture and Hyperparameters}
This section details the MEAP fine-tuning hyperparameters for the Llama3 model. The maximum sequence length is 4096 tokens, optimizing long-range dependencies and efficiency. The batch size is 512, and the learning rate schedule includes a warm-up for the first 10\% of training steps. The AdamW optimizer is used with $\beta_1 = 0.9$ and $\beta_2 = 0.95$, and the learning rate is set to $2 \times 10^{-5}$.

\begin{table*}[h]
    \caption{MEAP fine-tuning hyperparameters of Llama3 model}
    \vskip 0.1in
    \label{tab:llama_hyperparameters}
    \vskip 0.15in
    \begin{center}
    \begin{tabular}{cc}
    \toprule
    \textbf{Name} & \textbf{Hyperparameter} \\
    \midrule
    optimizer & AdamW \\
    lr schedule & cosine \\
    clip & 1.0 \\
    learning rate & $2 \times 10^{-5}$ \\
    weight\_decay & $5 \times 10^{-2}$ \\
    maximum sequence length & 4096 \\
    batch size & 512 \\
    \bottomrule
    \end{tabular}
    \end{center}
    \vskip -0.1in
\end{table*}

\end{document}